\documentclass[11pt]{article}

\usepackage[final]{acl}

\usepackage{times}
\usepackage{latexsym}
\usepackage{tabularx}
\usepackage{booktabs}
\usepackage[T1]{fontenc}

\usepackage[utf8]{inputenc}

\usepackage{microtype}

\usepackage{inconsolata}

\usepackage{graphicx}

\title{CaresAI at SMM4H-HeaRD 2026: Predicting TNM Staging}

\author{
\textbf{Joseph Itopa Abubakar}$^{1}$,
\textbf{Jorge Jarme}$^{1,2}$,
\textbf{Favour Igwezeke}$^{1,3}$,
\textbf{Mary Adewunmi}$^{1,4\dagger}$
\\
$^{1}$CaresAI, Australia \\
$^{2}$Ateneo De Naga University, Philippines \\
$^{3}$Faculty of Pharmaceutical Sciences, Nsukka, Enugu, Nigeria \\ 
$^{4}$Menzies School of Health Research, Australia\\
\\
\texttt{joseph.itopaa@gmail.com},
\texttt{jorjarme@gbox.adnu.edu.ph}, \\
\texttt{favour.igwezeke.249461@unn.edu.ng},
\texttt{mary.adewunmi@caresai.org}
}

\begin{document}
\maketitle
\begin{abstract}
The Tumor, Node, and Metastasis (TNM) staging system is critical to cancer treatment. This study aims to predict TNM stage labels independently, with the Cancer Genome Atlas (TCGA) pathology report as the sixth shared task of SMM4H-HeaRD 2026. The problem is framed as three multi-label classification tasks. We explore both classical and deep learning approaches using Term Frequency-Inverse Document Frequency (TF-IDF) features and embeddings from ClinicalBERT, BioBERT, and PubMedBERT. These representations are used with Logistic Regression (LR), Light Gradient Boosting Machine (LightGBM), Feed-Forward Neural Networks (FFNN), and Wide Residual Networks (WRN). Our results show that individual embeddings perform similarly to the TNM label classification, while their combination improves its predictive ability. WRN achieves AUROC scores of 0.839 (T), 0.8502 (N), and 0.803 (M) with F1-scores of 0.622, 0.702, and 0.9337, respectively, for the training phase. 
LightGBM with TF-IDF performs best with AUROC scores of 0.9368 (T), 0.9524 (N), and 0.8311 (M) and F1-scores of 0.7559 (T), 0.7384 (N), and 0.7017 (M) during the training phase. Furthermore, the result of the Codabench for the test sets indicates a Macro-F1 score of 0.978, 0.957, and 0.879 for the T, N, and M categories respectively for test set 1; while test set 2 records a Macro-F1 score for T, N, and M is 0.807, 0.767, 1.0 respectively.
However, performance declined during the evaluation phase of the test sets, a drop from 0.938 for test set 1 to 0.858 for test set 2, for the Macro-F1 score across all stages; suggesting limitations in model generalizability, sensitivity to class imbalance, and challenges in processing lengthy clinical documents. Although this study provides an efficient baseline model and a reproducible pipeline, further optimization and validation are required before it can be considered suitable for use in a real-world clinical setting. 

\textbf{Keywords:} TNM (Tumor, Node, and Metastasis), Wide Residual Network, Transformer Models, Natural Language Processing (NLP), Machine Learning
\end{abstract}

\section{Introduction}

Cancer remains a leading cause of death globally, with its burden varying across populations, regions, and types \citet{who2024cancer}. Effective cancer management depends on accurate prediction of its progression at the time of diagnosis. Central to this is cancer staging, the systematic evaluation of tumor size, anatomical location, and extent of spread from the primary site, which directly informs treatment decisions and prognostic outcomes. The TNM staging system jointly maintained by the American Joint Committee on Cancer (AJCC) and the Union for International Cancer Control (UICC) is a globally adopted framework that provides a standardised "common language" for clinicians, researchers, and cancer registries in classifying disease stage. The system classifies cancer progression into three dimensions: the size and local invasiveness of the primary tumor (T), the spread to lymph nodes (N) and the presence of distant metastasis (M)\citet{pan2024tnm9}. 

NLP methods have been widely applied to clinical text for cancer-related tasks, including TNM staging, diagnosis, and extraction of prognosis from pathology reports \citet{hands2025nlp}. For TNM staging specifically, previous work by Kefeli(2024) has established that 
transformer-based models trained on TCGA pathology reports 
can classify T, N, and M labels across multiple cancers 
types with strong generalization means the model 
performs well on data from  varying patients 
populations unseen during training, achieving an 
AU-ROC of 0.815--0.942 on an independent external 
validation set \citet{kefeli2024bbten}. TF-IDF remains a reliable baseline for text classification due to its simplicity and proven retrieval performance \citet{salton1988}, while LightGBM has demonstrated fast and accurate classification on feature-based inputs across large datasets \citet{ke2017lightgbm}. This study aims to predict T, N, and M staging classification independently of the TCGA pathology report to support consistent and accurate clinical decision-making in oncology as the sixth shared task of SMM4H-HeaRD 2026. Further, by leveraging machine learning, the proposed approach seeks to reduce variability in staging, which is critical for treatment selection, prognosis estimation, and patient management.

\section{System Overview}

This section outlines the stages involved in addressing the TNM staging classification. 

\subsection {Dataset}
The dataset is a CSV file extracted from an originally PDF file of a de-identified free-text pathology report from TCGA. These reports comprise several features, including patient file name, text, T, N and M, which vary in length, structure, and terminology (such as domain-specific abbreviations, numerical measurements, and narrative-style descriptions) \citet{kefeli2024tcgareports}. The dataset is moderately imbalanced across TNM categories. To handle missing or incomplete labels, we processed each TNM category independently. This is because a value that appears missing for one category (e.g., T) may still be valid or informative for another (such as N or M stages). As a result, the dataset was filtered separately for each classification task.

In the training set of 6,774 records, the M label is 
particularly skewed: M1 (metastasis present) accounts for 
only approximately 7\% of labeled M cases, while M0 
makes up the remaining 93\% \citet{kefeli2024bbten}. 
Similar imbalances exist across T and N subcategories, with advanced stages such as T4 and N3 occurring less frequently than earlier stages across the corpus. Additionally, reports may include irrelevant content or ambiguous phrasing, making robust natural language processing methods necessary for reliable prediction. The corpus is divided into training, validation, and test sets, with TNM labels obtained from the structured clinical metadata provided within TCGA.
The dataset contains several key features. Each record includes a unique patient identifier (file number), a text report, and the corresponding TNM labels. The patient file number is a unique string used to distinguish each case, while the text field contains the pathology report written for that patient. The TNM labels represent cancer staging. For modeling purposes, the T (Tumor) labels, originally given as 1 to 4, are adjusted to 0 to 3. The N (Node) labels range from 0 to 3, and the M (Metastasis) labels are binary, taking values of 0 or 1.
The dataset was divided into three subsets: (A) a training set containing 6,774 records, (B) a validation set containing 2,279 records, and (C) a test set containing 499 records.
The validation set includes the patient identifier and report text, while the test set contains a patient ID and the corresponding report. Before modeling, the text was preprocessed by removing stop words and noise, breaking the text into tokens (tokenization), and reducing words to their base form (lemmatization). After these steps, the processed text is converted into numerical representations using TF-IDF and autoencoder-based embedding techniques. 

\subsection {Methodology}
We adopted a hybrid methodology that combines traditional and deep learning neural network models, from data preprocessing to model development.

In the preprocessing stage, we utilized traditional feature weighting methods on the training data, particularly Term-Frequency–Inverse Document Frequency (TF-IDF), to capture important lexical patterns and words in the text. Furthermore, TF-IDF was used as a baseline method for comparison with neural network approaches, including embeddings generated from pretrained neural networks. Subsequently, we extracted and averaged the embedding output 
from three pretrained biomedical language models to generate 
dense text representations: ClinicalBERT 
\citet{huang2020clinicalbert}, BioBERT \citet{lee2020biobert}, 
and PubMedBERT \citet{gu2021pubmedbert}. These models generated dense embeddings that capture deeper semantic meaning in the training datasets than TF-IDF.

For the modeling stage, each pathology report in the 
training set is associated with structured TNM labels 
provided in the TCGA clinical metadata. These labels 
were not engineered by us, but were taken directly from 
the dataset as supplied for this shared task. The T 
(Tumor) labels were originally provided as values 1--4 
and adjusted to 0--3 for modeling purposes. The N (Node) 
labels range from 0 to 3, and the M (Metastasis) labels 
are binary, taking values of 0 or 1.

The traditional Machine Learning models used for training are as follows: 

A) Logistic regression - A Baseline Model that is used for classification tasks. It works well with both continuous and categorical data and is particularly useful for predicting discrete outcomes. In this study, the selected model performs well as a baseline and a base learner for the ensemble used, especially for the M category.

B)  Ensemble Methods of LightGBM and Random Forest - LightGBM is utilised for high-dimensional and sparse data, such as  TF-IDF, to improve performance, while Random Forest builds multiple decision trees and combines their output. LightGBM handles sparse data well and performs implicit feature selection. We also built a stacked ensemble, where random forest and LightGBM act as base models, and logistic regression acts as a meta-model to combine their predictions. This setup helps improve overall performance by taking advantage of the strengths of multiple models.

The advanced models used for training are the deep learning models, including Feed-Forward Neural Networks (FFNNs) and Wide Residual Neural Networks (WRNs).
FFNNs are a type of basic neural network where information flows in one direction, from input to output. They are widely used for classification and pattern recognition \citet{ZHOU202219}.

WRNs are an extension of neural networks with wider layers, allowing them to capture more complex patterns. They are designed to improve performance while maintaining efficient training \citet{zagoruyko2017wrn}. The combination of TF-IDF features, pretrained BERT embeddings, and both classical and deep learning 
classifiers allows us to compare predictive ability across different levels of model complexity and 
representation learning.

\section{Results and Discussion}

We achieved a micro-F1 score of 0.94 in the validation set and 0.926 across all stages in the official test set for the TNM classification task in SMM4H-HeaRD 2026 Task 6. The micro-F1 score is likely to decrease for the test set due to an unforeseen class imbalance in training. Some labels, such as M1 (metastasis present), are very rare (only about 7 per cent of the M category of the training data). 

The M (metastasis) prediction task was the hardest because the word being not clearly stated in the reports but must be inferred from the context. This requires a deeper clinical understanding of metastasis rather than simple keyword matching. 

The following two tables display two sets of results from the experimental results and the results after submission to the Codabench platform. The training set used was split into 80 per cent for training and 20 per cent for testing. This did not change throughout the experiments, as shown in Table \ref{tab:model_performance}, including both the classical machine learning methods and the deep learning method. Table \ref{tab:model_performance} demonstrates the best results from the series of experiments conducted with two data pre-processing techniques (TF-IDF and auto-encoders). Using TF-IDF features on the training data, the best-performing model for the M category was LightGBM, achieving an AUROC of 0.8311 and an F1-score of 0.7017 during the training phase. Furthermore, an AUROC of 0.9524 and an F1 score of 0.7384 were achieved for category N. Finally, for the TF-IDF method, an AUROC of 0.9368 and an F1 score of 0.7559 were achieved for category T.
We achieved an AUROC of 0.96 on the validation set compared to the AUROC score of 0.815–0.942 from the organizer’s baseline.

\begin{table}[t]
\centering
\small
\begin{tabularx}{\linewidth}{llccc}
\toprule
\textbf{Group} & \textbf{P-T} & \textbf{Model} & \textbf{AUROC} & \textbf{F1} \\
\midrule
M & TFIDF & LGBM & 0.8311 & 0.7017 \\
N & TFIDF & LGBM & 0.9524 & 0.7384 \\
T & TFIDF & LGBM & 0.9368 & 0.7559 \\
N & A-E   & WRN  & 0.8502 & 0.7020 \\
T & A-E   & WRN  & 0.8390 & 0.6220 \\
M & A-E   & WRN  & 0.8030 & 0.9337 \\
\bottomrule
\end{tabularx}
\caption{Model performance by category and preprocessing method.}
\label{tab:model_performance}

\vspace{0.3em}
\footnotesize
\textbf{Keys:}
P-T = preprocessing technique;
TFIDF = term frequency--inverse document frequency;
A-E = autoencoder;
WRN = wide recurrent network;
LGBM = LightGBM.
\end{table}

The results obtained using autoencoder-based representations (BioBERT, ClinicalBERT, and PubMedBERT embeddings) demonstrate that switching between individual embedding methods does not cause significant performance differences. However, combining these representations leads to noticeable improvements in its prediction ability. Using the WRN model, T, N and M achieved AUROC scores of 0.839, 0.8502, and 0.803 with the corresponding F1-scores of 0.622, 0.702 and 0.9337, respectively.  
The final submission to Codabench is based on the best-performing combination of model and preprocessing method, particularly LightGBM with TF-IDF features. Strong evaluation performance is shown in Table \ref{tab:model_performance}, where macro-averaged metrics indicate an F1-score of 0.978 for T, 0.957 for N, and 0.879 for category M in Test 1, while the evaluation of Test 2 records 0.807, 0.767, and 1.000 for category T, N, and M respectively.
\begin{table}[t]
\centering
\setlength{\tabcolsep}{6pt}
\renewcommand{\arraystretch}{1.1}

\begin{tabular}{lcc}
\hline
\textbf{F1 score} & \textbf{Test 1} & \textbf{Test 2} \\
\hline
T & 0.978 & 0.807 \\
N & 0.957 & 0.767 \\
M & 0.879 & 1.000 \\
\hline
\end{tabular}
\caption{TNM Extraction Results from Codabench}
\label{tab:tnm_results}
\end{table}

The model performs moderately in test set 2 (Table \ref{tab:tnm_results}), with lower macro-averaged scores (approximately 0.7–1.0), indicating unequal performance between classes, while macro-averaged scores in Test set-1 remain approximately 0.9.

Globally across all categories for the test set 1, the model achieves a high Macro-F1 score of 0.938, suggesting that it can accurately learn patterns from the available data. However, the performance drops to a Macro-F1 score of 0.858 in test set 2, indicating a limited generalization to unseen data. This gap is likely due to factors such as class imbalance (rare TNM labels like M1), overfitting to the training distribution, and variability of the same medical conditions across reports. The lower macro-averaged scores further confirm that the model struggles with minority classes, which are often the most clinically important. (see Table \ref{tab:tnm_results}).

\section{Limitations}
A key limitation of this study is the reliance on relatively small and unbalanced clinical text data, where rare TNM classes (particularly M1) are underrepresented, leading to biased model performance. Additionally, the use of standard BERT-based models with a 512-token limit results in the truncation of many pathological reports, causing the loss of important contextual information. This, combined with high-dimensional feature representations such as TF-IDF, increases the risk of overfitting and reduces generalizability to unseen data. This makes models more likely to be biased towards common label combinations within the training data and struggle with rare ones. Another limitation is that standard BERT-based models can only process up to 512 tokens, but most TCGA reports are longer than this. This means parts of the reports are cut off, leading to loss of important information and lower performance. Another limitation was overfitting with TF-IDF; the data are relatively small, causing models' inability to generalize well, particularly with high-dimensional features. We did not experiment with focal loss in the deep learning models, as our investigations were not yet exhaustive, and we considered this an area for future work. 

\section{Conclusion}
Our study provides a simple, reproducible, and computationally efficient pipeline that combines TF-IDF and BERT-based embeddings with traditional and deep learning models.  It also highlights effective approaches and where limitations exist, including data imbalance, long clinical documents, and model constraints, thereby complicating TNM staging prediction from pathology reports. Regarding the traditional machine learning approaches employed, we applied stratification and oversampling to address class imbalance; however, no significant performance differences were observed between models trained with and without oversampling.

This study can be effectively utilised as a baseline model, but not yet reliable for predicting TNM staging from pathology reports in real-world clinical settings. Robust generalizability is essential for models tasked with predicting TNM staging across diverse and previously unseen cancer data. Models that perform well only on familiar datasets but degrade on new cases risk producing inaccurate staging predictions, potentially compromising treatment decisions and patient outcomes in real-world clinical settings.
However, this approach is still valuable as the best-performing methods of our experiments, across all stages achieving an overall validation F1 score of 0.96; an F1 score of 0.938 for test set 1, and 0.858 for test set 2; Finally, an AUROC score up to 0.9524. These results demonstrate the potential of the selected approach to enhance TNM staging prediction, thus reducing clinician workload and enabling more efficient treatment planning and oncology registry processes. Further, the computational efficiency of the model with LGBM and TF-IDF makes it practical for large-scale processing of clinical reports, which can be integrated into pipelines for pre-screening, assisting clinicians, or prioritizing cancerous cases in real-world clinical settings. 

\section{Acknowledgement}
We would like to specifically thank our team at CaresAI, specifically Reem Abdel Salam, Leon Hamnett and Minh Khem, for their contribution towards the success of this study.

\section{Code Availability}
The project code documentation can be found on GitHub: \href{https://github.com/CaresAI-AU/Task-6---Predicting-TNM-Staging-from-TCGA-Pathology-Reports.git}{GitHub Repository}

The key hyperparameters for the entire project: \href{https://docs.google.com/spreadsheets/d/1ZQHYDGu3ExgQaLOE5pQCJXzmZVMaBV6eeZRuovsI/edit?usp=sharing}{Hyperparameters}

\bibliography{custom.bib}
\end{document}